\documentclass[acmtog]{acmart}

\acmSubmissionID{138}

\usepackage{booktabs} 

\usepackage{enumitem}
\usepackage{amsmath}
\usepackage{amsfonts}
\usepackage{multirow}
\usepackage{bigdelim}
\usepackage{color}
\usepackage{booktabs}
\usepackage{ifthen}
\usepackage{hyperref}
\usepackage{subcaption}
\usepackage{bbold}
\usepackage[noabbrev]{cleveref}
\usepackage{xcolor, pgf}
\usepackage{color, colortbl}
\usepackage[normalem]{ulem} 
\usepackage{placeins}
\usepackage{soul}
\usepackage[skins]{tcolorbox}

\tcbset{commonstyle/.style={boxrule=0pt,sharp corners,enhanced jigsaw,nobeforeafter,boxsep=0pt,left=\fboxsep,right=\fboxsep}}

\newtcolorbox{mycolorbox}[1][]{commonstyle,#1}

\usepackage{algorithm}
\usepackage{algorithmicx}
\usepackage{algpseudocode}
\algnewcommand\algorithmicinput{\textbf{Input:}}
\algnewcommand\INPUT{\item[\algorithmicinput]}
\algnewcommand\algorithmicoutput{\textbf{Output:}}
\algnewcommand\OUTPUT{\item[\algorithmicoutput]}
\algnewcommand\algorithmicforeach{\textbf{for each}}
\algtext*{EndFor}

\algdef{S}[FOR]{ForEach}[1]{\algorithmicforeach\ #1\ \algorithmicdo}
\algrenewcommand{\alglinenumber}[1]{\color{red!80!blue}\footnotesize#1:}

\algnewcommand\Func[2]{\textcolor{green}{#1}\textcolor{green}{(#2)}}
\algnewcommand\Insert[2]{Insert {#1} to #2.}
\algnewcommand\Input[1]{\State \textbf{Input: } #1}
\algnewcommand\Output[1]{\State \textbf{Output: } #1}

\newlength\myboxwidth
\definecolor{gray}{rgb}{0.5,0.5,0.5}
\definecolor{green}{rgb}{0, 0.6, 0}
\definecolor{orange}{rgb}{1, 0.5, 0}
\definecolor{mahogany}{rgb}{0.75, 0.25, 0.0}
\definecolor{purple}{rgb}{0.6, 0, 0.6}
\definecolor{darkgreen}{rgb}{0, 0.3, 0}
\definecolor{orange}{rgb}{1, 0.5, 0.}
\definecolor{lightblue}{rgb}{0.52, 0.75,0.91}
\definecolor{softgreen}{rgb}{0.66,0.87,0.74}
\definecolor{softred}{rgb}{0.96,0.71,0.69}

\newcommand{\ignore}[1]{}
\newcommand{\none}[1]{}
\newcommand{\com}[1]{}

\newcommand{\etal}{{\it{et~al.}}}
\newcommand{\ie}{i.e.,}
\newcommand{\eg}{e.g.,}

\DeclareMathOperator*{\argmax}{arg\,max}
\DeclareMathOperator*{\minimize}{minimize}

\setcopyright{acmcopyright}
\copyrightyear{2022}
\acmYear{2022}
\acmDOI{10.1145/1122445.1122456}

\acmJournal{TOG}
\acmVolume{38}
\acmNumber{4}
\acmYear{2021}
\acmMonth{7}

\begin{document}
\title{AutoPoly: Predicting a Polygonal Mesh Construction Sequence from a Silhouette Image}

\author{I-Chao Shen}
\orcid{1234-5678-9012-3456}
\affiliation{%
 \institution{The University of Tokyo}
 \country{Japan}
}
\author{Yu Ju Chen}
\affiliation{%
 \institution{Tencent America}
 \country{United States}
}
\author{Oliver van Kaick}
\affiliation{%
 \institution{Carleton University}
 \country{Canada}
}
\author{Takeo Igarashi}
\affiliation{%
 \institution{The University of Tokyo}
 \country{Japan}
}

\renewcommand{\shortauthors}{Shen, et al.}

\begin{abstract}
Polygonal modeling is a core task of content creation in Computer Graphics.
The complexity of modeling, in terms of the number and the order of operations and time required to execute them makes it challenging to learn and execute.
Our goal is to automatically derive a polygonal modeling sequence for a given target. 
Then, one can learn polygonal modeling by observing the resulting sequence and also expedite the modeling process by starting from the auto-generated result. 
As a starting point for building a system for 3D modeling in the future, we tackle the 2D shape modeling problem and present \textit{AutoPoly}, a hybrid method that generates a polygonal mesh construction sequence from a silhouette image.
The key idea of our method is the use of the Monte Carlo tree search (MCTS) algorithm and differentiable rendering to separately predict sequential topological actions and geometric actions.
Our hybrid method can alter topology, whereas the recently proposed inverse shape estimation methods using differentiable rendering can only handle a fixed topology.
Our novel reward function encourages MCTS to select topological actions that lead to a simpler shape without self-intersection.
We further designed two deep learning-based methods to improve the expansion and simulation steps in the MCTS search process:
an $n$-step ``future action prediction'' network (nFAP-Net) to generate candidates for potential topological actions, and a shape warping network (WarpNet) to predict polygonal shapes given the predicted rendered images and topological actions.
We demonstrate the efficiency of our method on 2D polygonal shapes of multiple man-made object categories.
\end{abstract}



\begin{teaserfigure}
  \includegraphics[width=\textwidth]{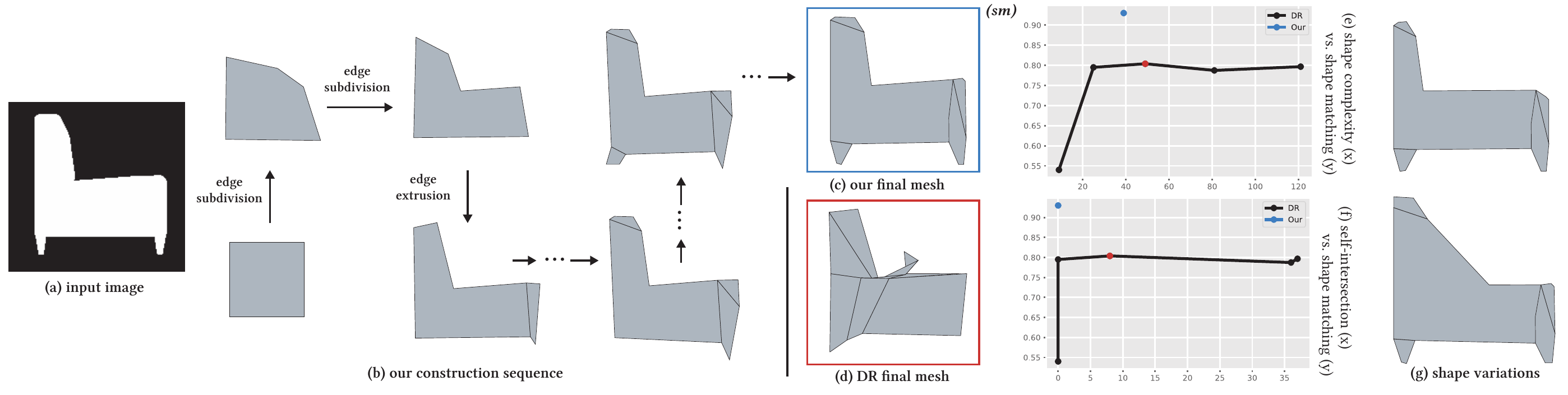}
  \caption{
  Given (a) an input silhouette image, our method generates (b) a mesh construction sequence leading to (c) our final mesh.
  Compared to (d), the result using inverse shape estimation with differentiable rendering (DR), our method  achieves a (e) higher shape matching score (y-axis) with less shape complexity and (f) fewer self-intersection (x-axis).
  In (e) and (f), the points in blue and red represent our and the DR final meshes, respectively.
  (g) Using the predicted construction sequence, user can easily create shape variations from the input image.
  }
  \label{fig:teaser}
\end{teaserfigure}

\maketitle

\section{Introduction}
\label{sec:intro}
Polygonal modeling plays the central role of visual content creation for various applications in Computer Graphics, such as game development, digital fabrication, and movie production.
These models are typically created by professional artists using polygonal modeling softwares, such as Maya~\cite{maya} and Blender~\cite{blender}.
However, the construction task remains tedious for professional artists, even for relatively simple shapes.
The main reason is that the entire construction process requires tens of thousands of repetitive actions, which are classified into four types: element selection, camera control, topological actions (\eg~subdivide face and extrude edge), and geometric actions (\eg~vertex translation and face rotation).

Novice users commonly learn how to perform polygonal modeling by following video or visualization tool-based tutorials detailing the construction sequence of their shape-of-interest.
However, it is difficult to closely follow every step when watching such tutorials without embedded hints~\cite{Lee:2011:SRU} or instructions.
The goal when modeling a specific shape is typically to emulate a real-world object, \ie~to replicate it in the digital world for various purposes.
However, it can be difficult to obtain the construction sequence for a particular object.
Therefore, a method to automatically generate unique construction sequences for specific objects is needed. 
The generated construction sequences can be used to construct interactive tutorials for teaching modeling or for easily creating variations of the object.

In this work, we present \textit{AutoPoly}, an algorithm for converting a silhouette image into valid polygonal mesh construction sequences.
Given a silhouette image of a desired shape (\Cref{fig:teaser}(a)), our method generates a construction sequence (\Cref{fig:teaser}(b)) for it from scratch.
Our method is inspired by a common practice of polygonal modeling called model-by-blueprint.
The modelling blueprint typically comprises 3-4 images showing orthogonal views of the desired shape. 
A modeller can trace a shape in two dimensions using the spatial information provided by the blueprint.

Although we can use differentiable rendering to obtain a mesh geometry that matches with target image~\cite{Li:2018:DMC, zhao2020physics}, there are still challenges of generating a construction {\em sequence} from images.
First, differentiable rendering method deform shape with fixed topology only~\cite{zhao2020physics}, \ie~we can not use it to predict topological actions. 
Second, generating a construction sequence can be tricky because applying one topological action might affect the subsequent topological and geometric actions of all shape elements.
It is laborious to search exhaustively for a feasible sequence of modeling actions because the search complexity is exponential.
These challenges apply even to a silhouette image.
Thus, we focus on generating a construction sequence for a 2D silhouette image first, which can already be used to create 2D shape variations or make 2D animations.
Solving the 2D problem will be an essential building block for generating a construction sequence for a 3D shape.

To address these challenges, we design \textit{AutoPoly} as a hybrid method that combines the Monte Carlo tree search (MCTS) algorithm~\cite{metropolis1949monte} and inverse shape estimation for predicting a mesh construction sequence for a silhouette image.
The main purpose of our method is to efficiently explore the possible topological action space using MCTS, and estimate the most likely shape given an updated topology using differentiable rendering.
We have also designed a novel reward function that promotes simpler geometry and avoids self-intersection.
However, although the final shape of our generated construction sequence matches the target shape, the method was highly inefficient because of the complex simulation step in the MCTS searching process.
To address this, we designed two deep learning-based methods: $n$-step ``future action prediction"" (nFAP-Net) and ``shape warping prediction'' (WarpNet).
The purpose of nFAP-Net is to predict possible future topological actions, the locations where these predicted actions will be applied, and the rendered results after future geometric actions.
However, since the output of the nFAP-Net is not sufficient to provide the future shape, we designed WarpNet to predict the warping function parameterized by the thin-plate spline (TPS) method.
We use nFAP-Net and WarpNet together to obtain an efficient expansion and simulation step during MCTS search.

We tested \textit{AutoPoly} on various types of man-made shapes to obtain feasible construction sequences for a target image.
We demonstrate that our method can discover topological actions useful for matching the topology of the shape (\eg~creating a hole).
Finally, we show that by combining MCTS with nFAP-Net and WarpNet, our method can discover feasible construction sequences that generate comparable final shapes in a fraction of the time.

\section{Related Work}
\label{sec:related}
Polygonal shape creation and editing is a fundamental problem in computer graphics.
In this section, we discuss previous works concerned with building polygonal shapes from different types of observations, and generating and utilizing polygonal mesh construction sequences.

\paragraph{Polygonal shape creation}
Polygonal meshes have long been reconstructed from still images, videos, depth scans, computed tomography, and magnetic resonance imaging scans in the fields of computer graphics and computer vision.
For some techniques, the first step is to convert the source measurements into oriented point clouds, which are subsequently  reconstructed as polygonal meshes using Poisson surface reconstruction methods~\cite{kazhdan2006poisson,screened}.
Other measurements are converted into volume fields and reconstructed to derive polygonal meshes using the marching cube algorithm~\cite{lorensen1987marching,chen2021nmc}.
Sketch-based modeling~\cite{rivers:3dmodeling,teddy,li2017bendsketch} is also being actively researched, and enables naive users to create polygonal shapes easily.
More recent works have focused on reconstructing 3D shapes from a single image.
These works often adopted the ``analysis-then-synthesis'' paradigm, \ie~they first analyzed the category of the target shape and then reconstructed the 3D shape through the deformation of component parts from existing 3D models~\cite{xu2011photo, huang2015single}. 
Another line of works involve manual interaction of pre-defined primitives to reconstruct a proxy shape from a single image~\cite{gingold_annotation,chen20133, shen2021clipflip}.

In recent years, many deep learning-based method reconstruct detailed man-made 3D shapes~\cite{wu2018learning, groueix2018papier,sun2018pix3d}, garment~\cite{zhu2020deep}, and architecture shapes~\cite{ren2021intuitive} from images.
Meanwhile, differentiable rendering has emerged as a tool for inverse polygonal shape estimation~\cite{zhao2020physics, nicolet2021large,Li:2018:DMC}.
These methods propagate derivatives through the image synthesis process, to minimize a desired objective function based on the shape parameters.

The aforementioned methods have several disadvantages.
First, they reconstruct the final polygonal shape without using a feasible construction sequence, which limits the potential for editing the resulting shapes.
Second, most deep learning-based and differentiable rendering methods can only generate shapes with fixed topologies, which limits the shapes that can be constructed.
Our method addresses these issues and aims to construct a shape through a construction sequence, including for non-fixed topologies.
The recovered construction sequences can easily be edited into other similar shapes.
\begin{figure*}[h!]
\centering
\includegraphics[width=0.95\linewidth]{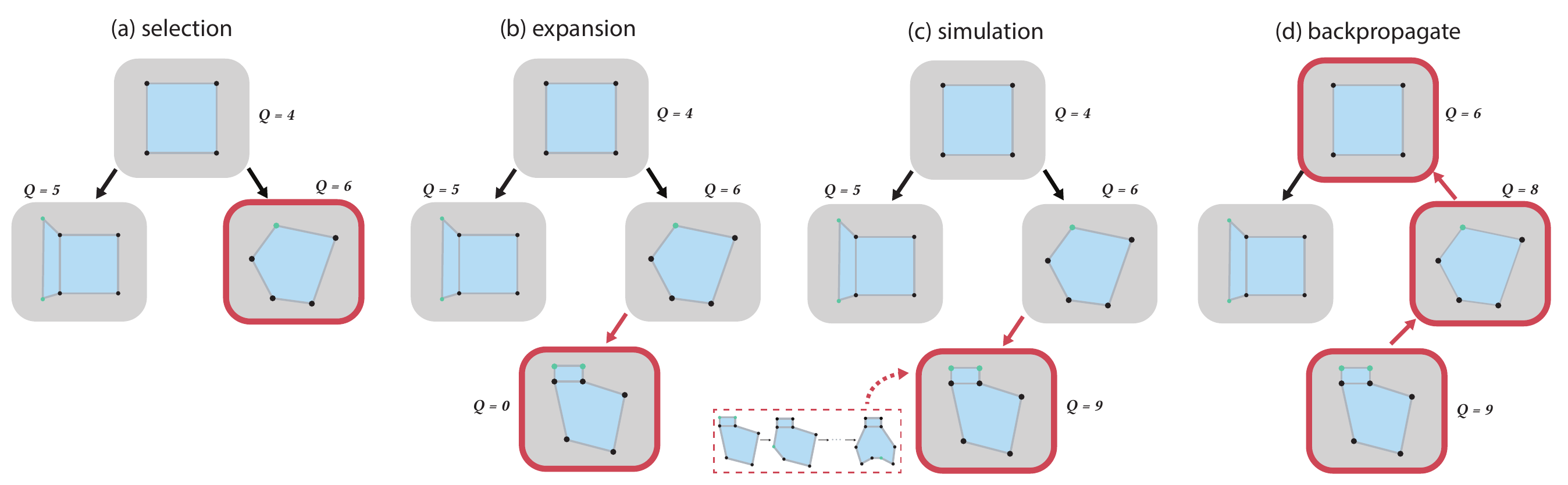}
\caption{
Illustration of the four steps of the Monte Carlo tree search process.
}
\label{fig:mcts_step}
\end{figure*}

\paragraph{Shape construction sequence and workflow}
Artists typically construct a shape using a step-by-step workflow.
Previous works have summarized the construction sequences for polygonal modeling~\cite{denning2011meshflow} and digital sculpting~\cite{denning20153dflow}.
These guidelines can be used as tutorials~\cite{denning2011meshflow} for polygonal modeling. 
Previous works have also analyzed the construction sequences for mesh version control~\cite{Denning:2013:MDM} and construction sequence retargetting~\cite{Salvati:2015:MCM}.
Du~\etal~\shortcite{du2018inversecsg} converted a polygonal mesh into a constructive solid geometry (CSG) representation.
Willis~\etal~\shortcite{willis2021fusion} introduced a parametric CAD model construction sequence dataset (\textit{Fusion360 gallery}) and a program synthesis method to generate the CAD construction sequence for a target shape.
Lin~\etal~\shortcite{lin2020modeling} proposed a two-step neural framework based reinforcement learning method model low resolution 3D shapes.
Our method focuses on recovering a feasible polygonal modeling construction sequence given a target image.
Unlike \cite{lin2020modeling}, our method do not need to train two separate reinforcement learning agents and we support more types of topological actions.
With our construction sequence, the user can generate a tutorial for a target shape~\cite{denning2011meshflow} and retarget the construction sequence into other shapes using \cite{Salvati:2015:MCM}.

\section{Method}
\subsection{Overview}
\begin{figure}[t!]
\centering
\includegraphics[width=0.9\linewidth]{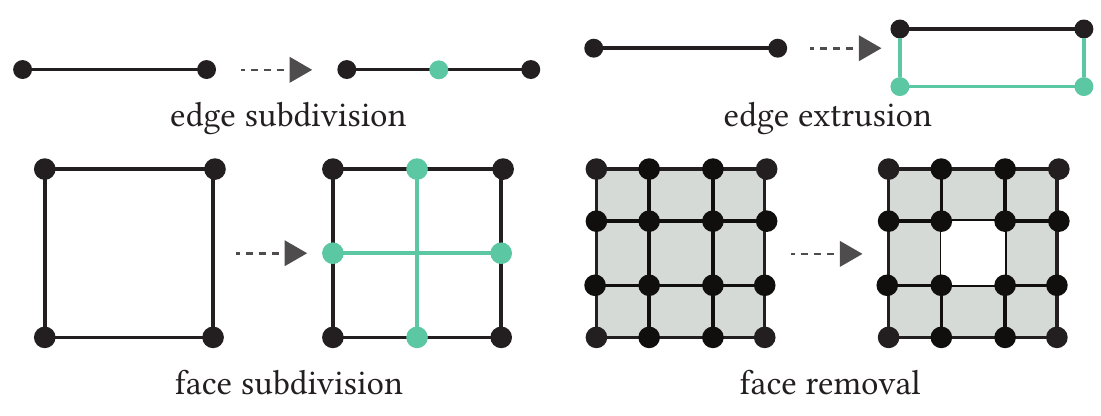}
\caption{
The four types of topological actions used by our method.
The black and green shape elements represent existing and updated elements, respectively.
}
\label{fig:topo_acts}
\end{figure}

Our goal is to obtain a feasible polygonal mesh construction sequence allowing the final shape to match the target shape.
The construction sequence involves both topological and geometrical actions.
A topological action is a modeling action that increases or decreases the number of shape elements, such as vertices, edges, and faces.
In this paper, we support four different kinds of topological actions as illustrated in \Cref{fig:topo_acts}.
By contrast, a geometric action transforms only the existing shape elements, such as by translation, scaling, or rotation.
The problem of recovering such a construction sequence can be viewed as a long-term planning problem because selecting each topological and geometric action affects other subsequent actions.
In this work, we propose a hybrid method that recovers sequential topological and geometric actions using MCTS~\cite{metropolis1949monte} and differentiable rendering for inverse shape estimation.
\begin{figure*}[!h]
\centering
\includegraphics[width=\linewidth]{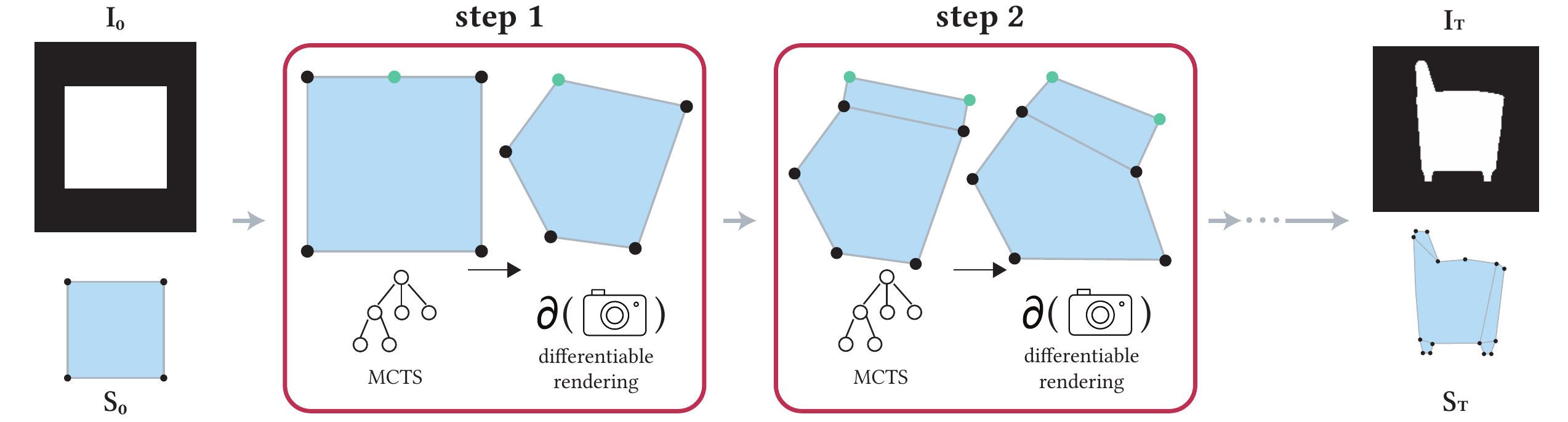}
\caption{
\textbf{Overview of our method.}
Given a target image $I_{T}$, we method iteratively perform steps comprise topological action search using MCTS and inverse shape estimation using differentiable rendering.
The resulted actions form the construction sequence of final shape $S_{T}$ that matches the shape in the target image $I_{T}$.
}
\label{fig:overview}
\end{figure*}
\subsection{Monte Carlo Tree Search}
Because the search space of the polygonal mesh construction sequence is extremely large, it is important to apply an efficient searching strategy for obtaining a feasible construction sequence from the input images.
Our method uses MCTS to search for feasible topological editing actions.
In each search iteration, MCTS constructs a new search tree, and performs random simulations based on various actions.
The simulation statistics of each action are stored to make future decisions more efficient.
After deciding on an action, MCTS constructs a new search tree and performs a similar search action iteratively until it fulfils the user-specified stopping criteria.

\paragraph{Search tree structure}
In a search tree, a node represents a shape, and a path in the tree represents a sequence of topological and geometric actions.
For example, by applying a topological action and a series of geometric actions, the shape represented by a parent node is transformed into the shape represented by a child node.
Each node stores a Q-value, which denotes to the expected future accumulated reward for matching the target shape, starting from the shape represented by that node.
\paragraph{Search iteration}
Each MCTS iteration consists of four steps as shown in \Cref{fig:mcts_step}:
\begin{enumerate}[label=(\alph*)]
    \item \textbf{Selection}: Selection starts from the root node ($N_{\text{root}}$) and, at each level, the next node is selected according to the the tree policy.
    The selection step terminates when a leaf node is visited and there is no valid topological action to be explored, or a terminal state of the environment has been reached.
    \item \textbf{Expansion}: If the shape of the selected node ($N_{\text{s}}$) does not match the target shape, a valid topological action is selected randomly.
    A new expanded node ($N_{\text{exp}}$) is created by applying the selected topological action to the shape of the selected node, and performing an inverse shape estimation using the target shape.
    \item \textbf{Simulation}: Perform a random simulation based on the shape represented by the expanded node. 
    During the simulation, a sequence of topological andgeometric actions are performed according to a desired policy, until either the maximum depth is reached or the current shape matches the target shape.
    After the simulation, the Q-value of $N_{\text{exp}}$ is updated according to the accumulated rewards associated with the simulation result.
    \item \textbf{Backpropagation}: The updated Q-value of $N_{\text{exp}}$ is back-propagated toward $N_{\text{root}}$.
\end{enumerate}

\subsection{Inverse shape estimation using differentiable rendering}
Given a shape with a fixed topology, the goal is to estimate the geometric actions (\ie~transformations of shape elements) needed to transform it to best match the input silhouette image the most.
To achieve this goal, recent works have adopted differentiable rendering to estimate the shape and material of an object from one or more input photographs.
The goal of differentiable rendering is to estimate the vector of scene parameters $\mathbf{p}$, including the shape geometry, materials, scene lighting, and camera parameters, given a target image $I^{T}$.
This process can be thought of as $\mathbf{p} = \mathcal{R}^{-1}(I^{T})$, where $\mathcal{R}$ is the rendering function.

In this work, we focus on the inverse shape estimation problem using differentiable rendering so our scene parameter only contains mesh vertex positions. 
More specifically, we only optimize the mesh vertex positions $\mathbf{x}$ using the following formulation:
\begin{align}
\minimize_{\mathbf{x}\in\mathbb{R}^{3n}} \quad \Psi(\mathcal{R}(x), I),
\label{eq:diff_render}
\end{align}
where $\Psi$ is a loss function measuring the difference between the reconstructed image $\mathcal{R}(x)$ and the target image $I^{T}$, and $n$ is the number of vertices; this formulation can be solved by applying the gradient-based iterative solver $\mathbf{x} \leftarrow \mathbf{x} - \eta \frac{\partial \Psi}{\partial \mathbf{x}}$, where $\eta > 0$ is the step size. 

\subsection{Hybrid search process}
\label{sec:search_process}
In \Cref{fig:overview}, we illustrate the overall search process used to generate a feasible construction sequences that match the target shape in $I_{T}$.
At each iteration $i$, starting from the root node representing the initial shape $S_i$, we create a new search tree.
After growing the current search tree, we select the topological action $t_i$ associated with the child node with the highest Q-value.
Next, we apply the selected topological action $t_i$ to $S_i$, perform an inverse shape estimation by optimizing \Cref{eq:diff_render} and obtain the final vertex positions $\mathbf{x}_{i+1}$.
To reduce the ambiguity in the possible geometric transformations leading to $\mathbf{x}_{i+1}$, we focus only on vertex translations.
Specifically, the geometric action we obtain is $g_i = \{(\Delta x_j, \Delta y_j, \Delta z_j) \}_{j=1}^{n}$, where $n$ is the number of vertices in $S_i$.
The shape represented by the next node is $S_{k+1}=S_k\odot t_k \odot g_k$.
After $k$ iterations, we will obtain a construction sequence $\mathit{A}=((t_0, g_0), (t_1, g_1),...,((t_k, g_k))$.
The process can be formulate as maximizing the accumulated reward for constructing the initial shape $S_0$ into the target shape $S_{T}$:
\begin{align}
\argmax_{\mathit{A}} \sum_{k=1}^{\|\mathit{A}\|} r(S_k\odot t_k \odot g_k),
\end{align}
where $r(\cdot)$ is a reward function.

\subsection{Reward function}
\label{sec:reward}
The reward function measures how closely the current shape matches the target shape.
Designing a reward function for the polygonal mesh construction is nontrivial associated with quantitatively defining construction progress and
balancing conflicting objectives, such as matching the target shape while preventing high shape complexity and self-intersection.
We propose a novel reward function:
\begin{align}
r_{\text{all}} = w_{sm}r_{sm} - w_{sc}r_{sc} - w_{si}r_{si},
\end{align}
where $r_{sm}$ is the shape matching reward, $r_{sc}$ is the shape complexity penalty, and $r_{si}$ rewards a shape without self-intersection. 
The contribution from each term is controlled by their respective weights, $w_{sm}$, $w_{sc}$, and $w_{si}$.
We used $w_{sm}=100, w_{si}=5$ during all experiments in this paper.
We determined $w_{sc}=1$ empirically for each shape category, using $w_{sc}=1$ and $0.3$ for synthetic and ShapeNet shapes, respectively.

Given a shape $S$ and a selected action pair $(t, g)$, we obtain the next shape ${S}' \leftarrow S\odot t \odot g$ using a polygonal modeling simulator (Blender~\cite{blender}) and its reward as, $r_{\text{all}}({S}')$.
We define each reward in $r_{\text{all}}$ as follow:
\begin{itemize}
    \item \textbf{Shape-matching reward}: the intersection over union (IoU) score between the binarized reconstructed image $(\widetilde{R({S}')})$ and the silhouette target image $I_{T}$:
    \begin{align}
    r_{{sm}} = \frac{|\widetilde{R({S}')} \cap {I_{T}}|}{|\widetilde{R({S}')} \cup {I_{T}}|}.
    \end{align}
    \item \textbf{Shape complexity penalty}: the summation of the shape element number:
    \begin{align}
    r_{gc} = (|V| + |E| + |F|)
    \end{align}
    \item \textbf{Self-intersection penalty}: the number of intersections in the shape: $r_{{si}}=|\text{self-intersection}|$.
\end{itemize}

\section{Learning-based method}
Although we can obtain a feasible construction sequence using the hybrid search process described in \Cref{sec:search_process}, it is too inefficient for complex shapes.
The main bottleneck is that in the hybrid search process, a huge number of inverse shape estimation problems (\ie~solve \Cref{eq:diff_render}) need to be solved in the expansion and simulation steps of MCTS.
To accelerate this part of the hybrid search process, we designed a learning-based method with two parts: \textit{future action prediction} and \textit{shape warping prediction}.
We designed the former part, nFAP-Net, to predict the future topological actions, the locations where these actions will be applied, and the rendered results in the subsequent $n$ steps.
We designed the latter part, WarpNet, to efficiently estimate the mesh shape that matches the predicted topological actions and predicted resulting images from nFAP-Net.

\subsection{Future action prediction}
\label{sec:nfepnet}
\subsubsection{Input and output}
Starting from step $0$, the input to nFAP-Net is the rendered image of the current shape ($I_0$) and the target shape image ($I_{T}$).
We stacked $I_0$ and $I_{T}$ as the input of the network.
The output of the network is an $n$-step construction sequence $\mathcal{E}_n=\{(\hat{I}_k, \hat{a}_k, \hat{I}^a_k )\}_{k=1}^n$, where $\hat{I}_k$ is the predicted rendered image of step $k$, $\hat{a}_k$ is the predicted topological action, and $\hat{I}_k^a$ is the predicted image of action region.
\begin{figure}[t!]
\centering
\includegraphics[width=\linewidth]{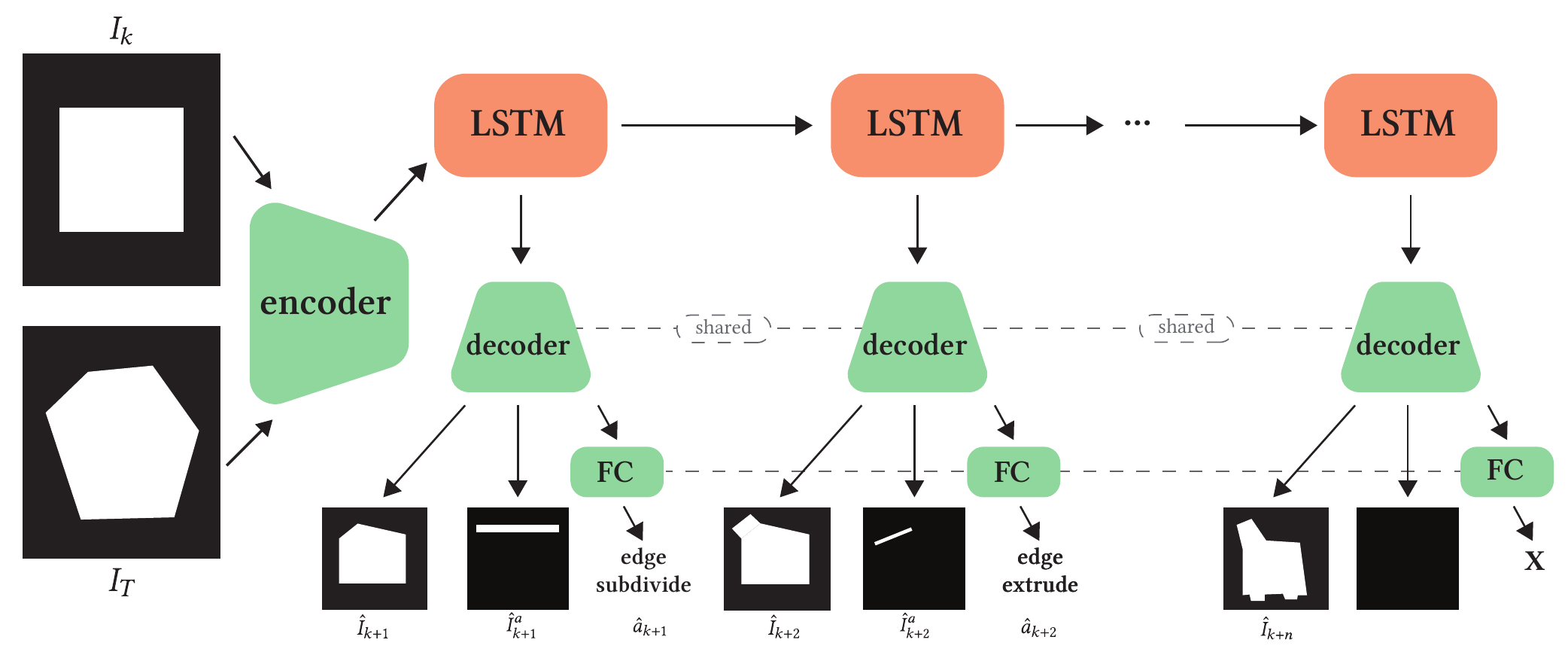}
\caption{
The architecture of our future action prediction network (nFAP-Net), where FC represents fully connected layer, and LTSM represents long short-term memory.
}
\label{fig:learn_arch}
\end{figure}

\subsubsection{Learning structure}
\Cref{fig:learn_arch} illustrates the structure of nFAP-Net.
At its core is a recurrent nueral network (RNN) that predicts a sequence of construction actions and corresponding rendered results.
Our reason for using this recurrent architecture is that it has proven effective for processing sequential data.
The specific architecture that we use employs Long Short-Term Memory (LSTM)~\cite{hochreiter1997long} components, which form a type of temporal memory controlled by special gates connected to its inputs.

In nFAP-Net, we have $n$ subnetworks (each column in \Cref{fig:learn_arch}), which predict topological actions and rendered shape image for the future $n$ step.
Before all the subnetworks, we use a CNN encoder to create an $\mathbb{R}^{128}$ feature vector for the input stacked image. 
This feature vector is then fed to an LSTM that learns the
relationships between the features, the future construction actions and rendered results.
Specifically, we set the initial state of the LSTM to a vector of zeros. 
After feeding the feature and zero vectors into the LSTM, the unit returns the next state and another output, which represents future actions and the rendered results.
A CNN-based decoder and set of fully-connected layers then decode the output of the LSTM into the predicted rendered image, topological action, and action region image. 
Then, the next stage is provided, along with the feature vector as an input to the LSTM again, to obtain the subsequent construction action.
We repeat this procedure $n$ times to obtain the $n$-step construction sequence $\mathcal{E}_n$.





\subsubsection{Training and losses}
nFAP-Net predicts multiple types of output simultaneously and thus requires a suitable combination of loss functions:
\begin{align}
\mathcal{L}(\mathcal{E}, I_{T})=(\mathcal{E})= \sum_{k=1}^{n} (\|I^T_{k} - \hat{I}_k\|_2^2 + CE(a_k, \hat{a}_k) +  \|I^a_{k} - \hat{I}^a_k\|_2^2),
\end{align}
where $CE$ is the cross entropy loss function, $I^T_k$ is the ground truth rendered image of the shape at step $k$, $a_k$ is the ground truth topological action applied at step $k$, and $I^a_{k}$ is the ground truth action region image at step $k$.

\subsubsection{Training data}
\label{sec:train_data}
To train our nFAP-Net, we needed training data with suitable construction sequences for different polygonal meshes.
To generate these data, we designed a random shape construction simulator.
For the random construction process, we start from a rectangle with four vertices, four edges, and one face.
We then randomly select a geometric element (vertex, edge, or face) and apply a random topological action to it.
Then, we apply random translations to all vertices on the shape.
This random construction process is illustrated in the supplemental material.

\subsection{Shape warping prediction}
\label{sec:warpnet}
The predicted output of the nFAP-Net $\mathcal{E}_n$ is inadequate for supporting both the expansion and simulation steps.
This is mainly because the predicted future rendered images ($\hat{\mathbf{I}}$), topological actions ($\hat{\mathbf{a}}$) and action region images ($\hat{\mathbf{I}^a}$) in $\mathcal{E}$ do not provide us with the future shapes ($\hat{\mathbf{s}}$).
This leads to two problems in terms of supporting MCTS steps.
First, we cannot compute the self-intersection penalty without the future shapes ($\hat{\mathbf{s}}$).
Second, and more importantly, we need the $1$-step future shape, performed after any topological and geometric actions,  $s_{k+1}$, to represent the expanded node.
To address these issues, one possible approach is to solve iterative differentiable rendering problems (\Cref{eq:diff_render}) to obtain the future shapes.
However, this will increase the computational cost again, thus thwarting the original purposes of designing learning-based method.
Instead, we designed WarpNet to predict the geometric action that matches the next predicted shape image.

The input to the WarpNet is the rendered image of the shape at step $k$ ($I_k$) and the target shape image ($I_{T}$).
Our goal is to warp $I_k$ to match $I_{T}$, and compute the underlying shape using the warped image.
To do this, we define an $M\times M$ grid $G_k$ on top of $I_k$ and use a thin-plate spline (TPS)~\cite{courant89} transformation as our warping function.
Such a transformation consists of an affine transformation $\mathcal{A}\in\mathbb{R^{2\times 3}}$ and a deformation field $\mathcal{D}\in \mathbb{R}^{M\times M \times 2}$.
The output of the WarpNet is the parameters of the TPS transformation: $\theta=(\mathcal{A}, \mathcal{D})$.
The warped image can be represented as $\mathcal{S}(\mathcal{T}_\theta(G_k))$, where $\mathcal{S}$ a differentiable bilinear sampler.
We use the following loss function to train the WarpNet:
\begin{align}
\mathcal{L} = \|\mathcal{S}(\mathcal{T}_\theta(G_k))-I_{T}\|_2.
\end{align}


At inference time, we use the predicted $\hat{\theta}$ to warp $G_k$ into $\hat{G}_k$.
For each vertex $v_q$ in $s_k$, we compute its new vertex position with $\hat{G}_k$ using the barycentric coordinates.
First, we find the face $\mathit{f}[v_q]$ in $G_k$ that contains $v_q$ and compute its barycentric coordinates $(\alpha_k^q, \beta_k^q, \gamma_k^q)$.
We then compute the new position of $v_q$ as: $\hat{v}_q=\alpha_k^q a+\beta_k^q b + \gamma_k^q c$, where $(a,b,c)$ is the vertex position of the corresponding face $\mathit{f}[v_q]$ in $\hat{G}^k$.

\subsection{MCTS with nFAP-Net and WarpNet}
In \Cref{alg:expansion}, we summarize how we use the trained nFAP-Net and WarpNet to support an efficient MCTS search in the expansion and simulation steps .

\begin{algorithm}
\caption{Expansion and simulation steps 
$n$-step future action prediction (nFAP-Net) and shape warping prediction (WarpNet).}
\label{alg:expansion}
\begin{algorithmic}[1]
\INPUT $I_j = \mathcal{R}(S_j)$, $I_{T}$
\OUTPUT $n$-step future shape $\{S_{j+n}\}_{1}^{n}$, $n=1$ for expansion step and $N_{\text{sim}}$ for simulation step
\State $\hat{\mathcal{E}}_n \gets \text{nFAP-Net}((I_j, I_T))$ \Comment{\Cref{sec:nfepnet}}
\State $\hat{\theta}_j=(\hat{\mathcal{A}}_j,\hat{\mathcal{D}}_j) \gets \text{WarpNet}((I_j, I_T))$ \Comment{\Cref{sec:warpnet}}
\State $\hat{G}_j \gets \mathcal{T}_{\hat{\theta}_j}(G_j)$ \Comment{TPS transformation}
\State $S_{j+1} \gets \text{Warp}(S_{j},\hat{G}_j)$    
\State \Return $S_{j+1}$
\end{algorithmic}
\end{algorithm}

\begin{table}[ht!]
\centering
\caption{
Ablation studies and quantitative evaluation using synthetic shape data.
For $r_{sm}$ and $r_{all}$, the higher the better, and for $r_{sc}$ and $r_{si}$, the lower the better.
}
\begin{tabular}[t]{lcccc}
\toprule
Method & $r_{sm} \uparrow$ & $r_{sc} \downarrow$ & $r_{si}\downarrow$ & $r_{\text{all}}\uparrow$\\
\midrule
\small w/o shape complexity (\textcolor{blue}{\Cref{sec:sc_ablation}})  & 0.935 & 17 & 1.3 & 70\\
\small w/o self-intersection (\textcolor{blue}{\Cref{sec:si_ablation}}) & 0.946 & 13.6 & 2.1 & 70.5\\
\midrule
\rowcolor{lightblue!50}
\small AutoPoly (w/o learning)  & \textbf{0.964} & 12.1 & 0.7 & \textbf{80.8}\\
\midrule
\small DR (simple) & 0.732 & \textbf{9.0} & \textbf{0.0} & 62.2\\
\small DR  (complex) & 0.853 & 49.0 & 3.7 & 21.8\\
\rowcolor{lightblue!50}
\small AutoPoly (w/ learning) & 0.958 & 12.4 & 1.1 & 77.9\\
\bottomrule
\end{tabular}
\label{tab:abl_and_shape_quan}
\end{table}%
\section{Experiment and Results}
\subsection{Implementation details}
We used Blender~\cite{blender} as our polygonal modeling environment, implemented nFAP-Net and WarpNet in PyTorch~\cite{NEURIPS2019_9015}, and used redner~\cite{Li:2018:DMC} as our differentiable renderer.
We ran these experiments on a machine with an AMD Threadripper 1950X CPU, 64GB of RAM, and a GeForce 2080Ti GPU.
\subsection{Ablation study}
To verify the performance on our method, we performed two ablation studies of the reward function for MCTS, focusing on the quality of the resulting topologies.
To compare different versions of our method, we generated a synthetic shape dataset with $50$ shapes using the data generation process described in \Cref{sec:train_data}.
\subsubsection{Shape complexity ($r_{sc}$)}
\label{sec:sc_ablation}
To demonstrate the importance of the shape complexity penalty, we removed $r_{sc}$ from our reward function and ran MCTS using the random test dataset.
In \Cref{tab:abl_and_shape_quan}, note that without $r_{sc}$, the shape complexity increase significantly.
Meanwhile, the lower $r_{sm}$ suggests that the extra shape elements were redudant and ineffective for matching the target shape.
\subsubsection{Self-intersection ($r_{si}$)}
\label{sec:si_ablation}
To verify the usefulness of the self-intersection penalty, we ran MCTS using the reward function $r_{\text{all}}=w_{sm}r_{sm}-w_{sc}r_{sc}$ and the synthetic dataset.
In \Cref{tab:abl_and_shape_quan}, note that the self-intersection count without $r_{si}$ increased significantly.
Also note that $r_{sm}$ decreased at the same time, and that the resulting defective shapes with many self-intersections are not ideal for further editing.

\begin{figure*}[t!]
\centering
\includegraphics[width=\linewidth]{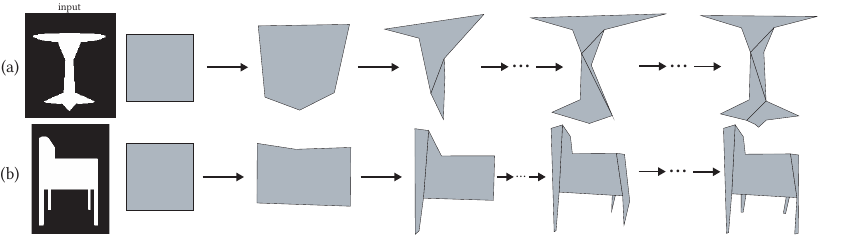}
\caption{
\textbf{Construction sequences derived using \textit{AutoPoly}}.
Given (a) an input silhouette image, \textit{AutoPoly} generates a 2D mesh construction sequence that matches the shape in the input silhouette image.
}
\label{fig:shapenet_seq}
\end{figure*}
\begin{figure}[t!]
\centering
\includegraphics[width=\linewidth]{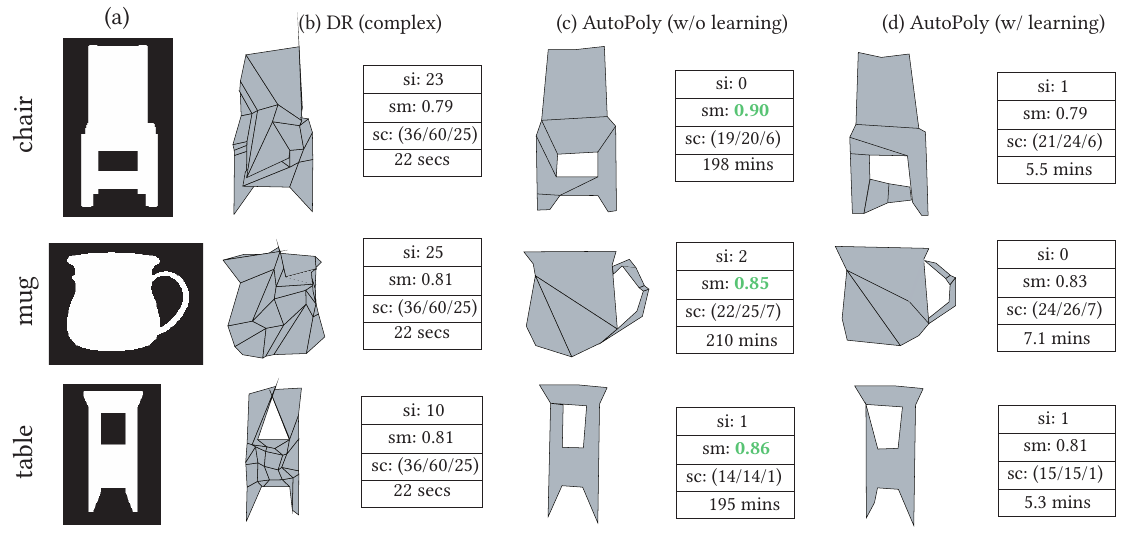}
\caption{
\textbf{Qualitative evaluation after reconstructing ShapeNet shapes.}
Compared to the pure differentiable rendering method using the \textit{complex} shapes, \textit{AutoPoly} generates final shapes that better match the target shape (sm), with a less complex shape (sc) and fewer self-intersections (si).
}
\label{fig:small_qual_comp}
\end{figure}
\subsection{Results of modeling sequence derivation}
\subsubsection{Quantitative evaluation on synthetic shape reconstruction.}
\label{sec:quan_eval}
We performed a quantitative evaluation of the 2D shape reconstruction on a synthetic shape dataset.
We compared our original method, the learning-based version of our method, and pure inverse shape estimation using differentiable rendering~\cite{Li:2018:DMC}.
For pure inverse shape estimation, we used two different initial shapes: (i) \textit{simple} shape: a rectangle with $4$ vertices, $4$ edges, and $1$ face, and (ii) \textit{complex} shape: a subdivided rectangle with $16$ vertices, $24$ edges, and $9$ faces.
We computed the mean shape-matching score, shape complexity score, and number of self-intersections, as listed in the bottom part (3rd - 6th row) of \Cref{tab:abl_and_shape_quan}; it can be seens that, although the inverse shape estimation using the complex shape obtains a higher shape-matching score given te higher degree-of-freedom, it also introduces greater shape complexity and a higher self-intersection count.
Our original method and the learning-based version obtained similar shape-matching scores, but with reduced shape complexity and a lower self-intersection count.
This suggests that our method can avoid complicating the shape with respect to the target shape, thus introducing fewer geometrical issues.
The ability to generate a construction sequence without obvious geometric defects is important for further editing.
The average time for generating a construction sequence using the ``AutoPoly (w/o learning)'' and ``AutoPoly (w/ learning)'' methods was $62$ minutes and $2.7$ minutes, respectively.
According to the performance indicators shown in \Cref{tab:abl_and_shape_quan}, the ``Full learn'' method can achieve comparable results in a fraction of the time.

\paragraph{Qualitative evaluation based on ShapeNet data}
In \Cref{fig:small_qual_comp}, we show the qualitative results of shape reconstruction for different shapes in ShapeNet~\cite{chang2015shapenet}.
We compared the shape reconstruction results among pure inverse shape estimation using differentiable rendering~\cite{Li:2018:DMC} for 200 iterations, \textit{AutoPoly} without learning, and \textit{AutoPoly} with learning.
Simialr to the results of the quantitative evaluation (\Cref{sec:quan_eval}), we can see that inverse shape estimation with the complex shape (\Cref{fig:small_qual_comp}(b)) introduced many self-intersections and increased shape complexity.
\textit{AutoPoly} with and without learning reconstructed the shape more efficient, with reduced shape complexity and a lower self-intersection count.
More importantly, \textit{AutoPoly} could derive a construction sequence that updates the topology of the original shape as shown in \Cref{fig:shapenet_seq}.
Note that in \Cref{fig:shapenet_seq}(a), \textit{AutoPoly} gradually generates the top of the table and then the table base.
In \Cref{fig:shapenet_seq}(b), \textit{AutoPoly} generates the seat first, and gradually generate the chair legs.

\section{Limitations and future work}
\paragraph{Derivation of 3D modeling sequence}
Our current method can only use a silhouette image as input and derive a construction sequence of a 2D mesh.
One main problem is the MCTS searching time in the 3D modeling search space is extremely high, so we cannot obtain a feasible construction sequence using the current method.
It is possible to design a reinforcement learning method to estimate the potential future reward that better guides the searching direction. 
\paragraph{Human-like construction sequence}
While the current reward function achieves a simpler mesh and avoids self-intersection, the resulting construction sequence does not match a how would construct the target shape.
In the future, 3D artists' construction sequences could be used as references, and a reward function could be designed to attempt to mimic these sequences.
\paragraph{User guidance}
Although our current method generates a feasible construction sequence, it sometimes focuses on regions that are not important for emulating the target shape.
It is possible to design a human-in-the-loop method to enable a user to provide guidance during the search process.
\section{Conclusion}
In this paper, we proposed a hybrid method for deriving a construction sequence that matches a target silhouette image.
We used MCTS and differentiable rendering to identify and select the suitable topological and geometric actions.
Our method generates a construction sequence that updates its topology (\eg~adds new shape elements) and avoids self-intersection.
We believe solving this problem is an essential building block for building a method for 3D modeling sequence derivation in the future.

\bibliographystyle{ACM-Reference-Format}
\bibliography{paper}

\end{document}